\documentclass[letterpaper, 10 pt, conference]{ieeeconf}
\usepackage{graphicx}
\usepackage{booktabs}
\usepackage{multirow}
\usepackage{amssymb}
\usepackage{amsmath}
\usepackage{subfigure}

\IEEEoverridecommandlockouts 
\title{\LARGE \bf
Unsupervised Monocular Depth Estimation Based on Hierarchical Feature-Guided Diffusion
}

\author{Runze Liu$^{1,2}$, Dongchen Zhu$^{2,3}$, Guanghui Zhang$^{2}$, Yue Xu$^{2}$, Wenjun Shi$^{2}$, \\ Xiaolin Zhang$^{1,2,3,4,5}$, Lei Wang$^{2,3}$, and Jiamao Li$^{2,3,*}$%
\thanks{*National Science and Technology Major Project from Minister of Science and Technology, China(2018AAA0103100), National Natural Science Foundation of China(62303441), Natural Science Foundation of Shanghai(23ZR1474200), Youth Innovation Promotion Association, Chinese Academy of Sciences(2021233, 2023242), Shanghai Academic Research Leader(22XD1424500)}%
\thanks{$^{1}$School of Information Science and Technology, ShanghaiTech University, Shanghai 201210, China.}%
\thanks{$^{2}$Bionic Vision System Laboratory, State Key Laboratory of Transducer Technology, Shanghai Institute of Microsystem and Information Technology, Chinese Academy of Sciences, Shanghai 200050, China.{$^{*}$Corresponding author: Jiamao Li (email:\tt\small jmli@mail.sim.ac.cn)}}%
\thanks{$^{3}$University of Chinese Academy of Sciences, Beijing 100049, China.}%
\thanks{$^{4}$Xiongan Institute of Innovation, Xiongan, 071700, China}%
\thanks{$^{5}$University of Science and Technology of China, Hefei, Anhui, 230027, China}%
}

\begin{document}

\maketitle

\thispagestyle{empty}
\pagestyle{empty}

\begin{abstract}
    Unsupervised monocular depth estimation has received widespread attention because of its capability to train without ground truth. In real-world scenarios, the images may be blurry or noisy due to the influence of weather conditions and inherent limitations of the camera. Therefore, it is particularly important to develop a robust depth estimation model. Benefiting from the training strategies of generative networks, generative-based methods often exhibit enhanced robustness. In light of this, we employ a well-converging diffusion model among generative networks for unsupervised monocular depth estimation. Additionally, we propose a hierarchical feature-guided denoising module. This model significantly enriches the model's capacity for learning and interpreting depth distribution by fully leveraging image features to guide the denoising process. Furthermore, we explore the implicit depth within reprojection and design an implicit depth consistency loss. This loss function serves to enhance the performance of the model and ensure the scale consistency of depth within a video sequence. We conduct experiments on the KITTI, Make3D, and our self-collected SIMIT datasets. The results indicate that our approach stands out among generative-based models, while also showcasing remarkable robustness.
\end{abstract}
    
\section{Introduction}
Monocular depth estimation aims to predict pixel-level depth and plays a crucial role in numerous applications such as autonomous driving, virtual reality (VR), and augmented reality (AR). With the rapid development of computer vision and deep learning, Eigen et al. \cite{supervised-depth} pioneer the application of deep learning to this field through a supervised approach. To reduce the model's data dependence, Zhou et al. \cite{sfmlearner} propose the first unsupervised framework for monocular depth estimation. Numerous works have optimized and improved depth estimation methods based on this initial framework \cite{sc-sfm, monodepth2, sc-depth, gan_vo, masked_gan, mc_gan}. These methods can be categorized into discriminative-based and generative-based methods, depending on their data modeling techniques through deep learning.

Discriminative-based monocular depth estimation methods \cite{sc-sfm, monodepth2, sc-depth} aim to learn the mapping from images to depth by maximizing the conditional probability distribution. These methods demonstrate impressive performance in ideally clear and high-quality images which are similar to the training set. However, in real-world scenarios, images captured by cameras may be affected by the weather conditions and the status of cameras. These will cause images in the test set blurry or noisy. Variations in data distribution between the test and training sets directly affect the mapping derived from the model, leading to poor robustness and failure in such scenarios. There are methods trying to improve the robustness of discriminative-based methods by adding perturbations to the training set \cite{diss}. In practical applications, the perturbations are diverse, including but not limited to illumination changes, blur, etc. These methods do not essentially improve the robustness of the model and still fail to handle scenarios that do not appear in the training set.

In contrast, generative-based monocular depth estimation methods \cite{gan_vo, masked_gan, mc_gan} could interpret the intrinsic distribution of depth by learning the joint probability distribution between images and depth. This approach exhibits greater robustness and adaptability when faced with novel data samples. Even when the input image is perturbed, such as the aforementioned scenarios, the model provides more accurate and robust depth estimation benefiting from the understanding of image and depth distribution. Kaneko et al. \cite{gan_denoise} demonstrate the strong robustness of generative networks when dealing with noisy images. In this work, we aim to continue to explore the application of generative networks in depth estimation and develop a robust unsupervised monocular depth estimation method.

\begin{figure}[t]\centering
    \includegraphics[width=0.47\textwidth]{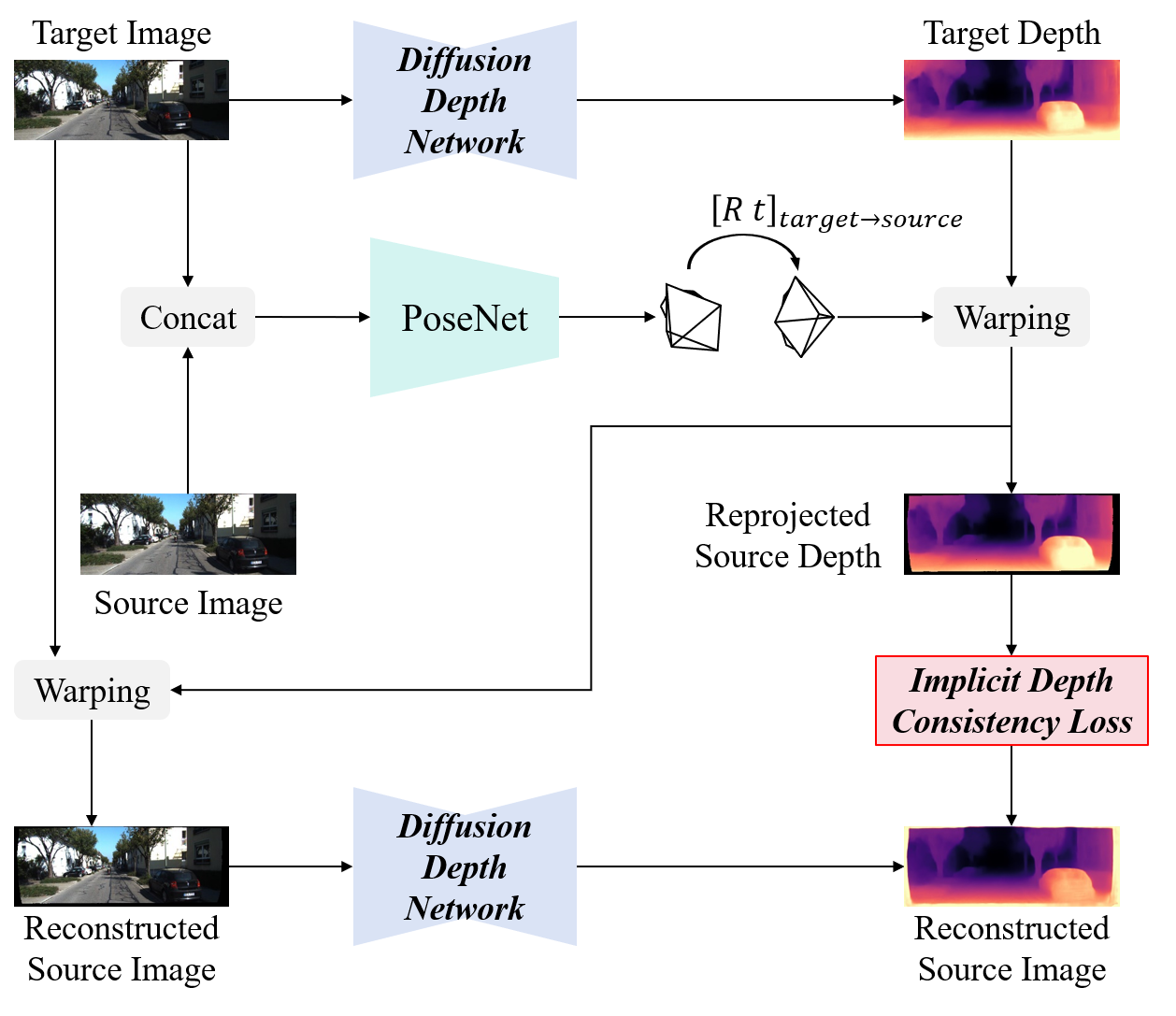}
    \caption{The framework of our proposed unsupervised monocular depth estimation based on the diffusion model. This framework consists of a depth estimation subnetwork and a pose estimation subnetwork. We integrate the diffusion model into the depth estimation subnetwork.}
    \label{framework}
\end{figure}
\begin{figure*}[!t]\centering
    \includegraphics[width=0.9\textwidth]{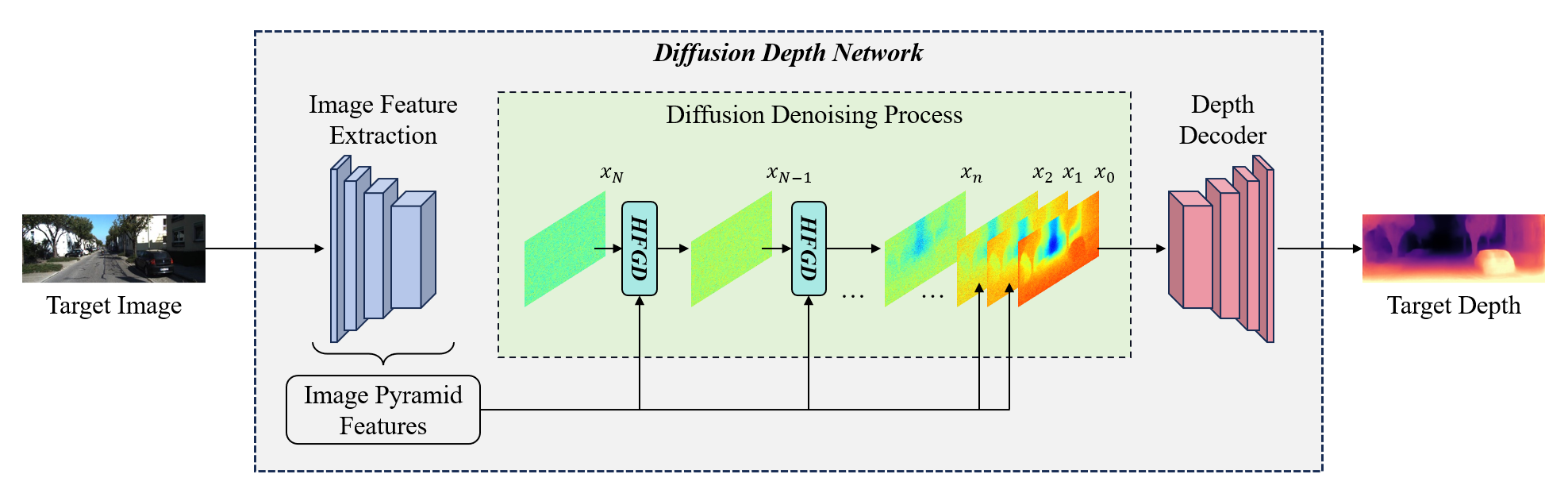}
    \caption{Illustration of the diffusion depth network in our proposed framework. Here `HFGD' stands for the `Hierarchical Feature-Guided Denoising Module'. The diffusion depth network aims to utilize image pyramid features to guide the denoising process. Image depth is finally obtained by denoising a random distribution.}
    \label{DDN}
\end{figure*}

Inspired by the successful application of a well-converging generative-based diffusion model \cite{ddim} in image feature enhancement \cite{dif_enhance} and panoptic segmentation \cite{dif_seg}, we propose an unsupervised monocular depth estimation framework based on the diffusion model, as shown in Fig.\ref{framework}. In this framework, we design a diffusion depth network by integrating the diffusion model into the depth estimation subnetwork, as illustrated in Fig.\ref{DDN}. The diffusion depth network iteratively refines a random distribution via a denoising process guided by an image, ultimately recovering depth from the random distribution. To enhance the model's capacity to learn and interpret the joint distribution of depth under image guidance, we propose a novel hierarchical feature-guided denoising module (HFGD), as illustrated in Fig.\ref{HFGD}. As we gradually integrate image pyramid features into each level of the denoising network, the guidance information evolves from low-level spatial geometric features to high-level semantic features. This approach allows for a more comprehensive utilization of the image, enhancing the model's interpretation of the depth feature distribution.

To constrain the depth estimation network effectively and enhance model performance, we propose an implicit depth consistency loss. During the training process, we fully explore the implicit depth information in reprojection. We utilize the depth of the source image obtained via reprojection as an implicit pseudo-label, aiming to constrain the depth of the reconstructed source image estimated by the network, as shown in Fig.\ref{framework}. The utilization of implicit deep consistency loss can more effectively constrain the depth estimation subnetwork within the model, thereby improving the depth prediction accuracy. Additionally, depth consistency across different frames ensures that the depths estimated by the model are consistent in scale within the same video sequence.

In summary, our contributions are as follows:
\begin{itemize}
    \item We propose a novel unsupervised monocular depth estimation framework based on the diffusion model, which exhibits strong robustness and demonstrates outstanding performance in complex scenes.
    \item We present a hierarchical feature-guided denoising module to fully utilize image pyramid features, which enables the model with a superior capacity to learn and interpret the depth distribution.
    \item We design an implicit depth consistency loss, which can better constrain the depth estimation subnetwork to enhance its performance and ensure the estimated depth at the same scale within a video sequence.
\end{itemize}

\section{Related Work}
\textbf{Unsupervised Monocular Depth Estimation based on discriminative networks.}
The foundational framework of unsupervised monocular depth estimation is first proposed by Zhou et al. \cite{sfmlearner}, which regards the depth estimation as the image generation from different views. This framework comprises a depth estimation subnetwork and a pose estimation subnetwork, trained through the optimization of reprojection photometric loss. Due to significant errors in photometric loss under varying environmental illumination, the structure similarity index measure (SSIM) \cite{ssim} is utilized to formulate a new reprojection loss \cite{ssim_use}. In scenarios where occlusions and dynamic objects invalidate the assumption of photometric consistency, Godard et al. \cite{monodepth2} introduce an automatic mask and a minimum reprojection loss mechanism to address these challenges. Considering the issue of monocular vision lacking an absolute scale, Bian et al. \cite{sc-sfm, sc-depth} propose a geometric consistency loss to constrain the estimated-depth remains consistent in scale. However, the effectiveness of this loss is compromised by the low performance of the pose estimation subnetwork during the early stages of training.

\textbf{Unsupervised Monocular Depth Estimation based on generative networks.}
Building upon the concept of generative networks, Almalioglu et al. \cite{gan_vo}  propose the first unsupervised monocular depth estimation framework based on generative adversarial network (GAN). This method enhances model robustness by generating depth with a generator and using a discriminator to constrain the difference between reconstructed and real images. Li et al. \cite{gan_vo2} further improve model robustness by employing the generator to generate both the depth and pose directly. To mitigate the influence of occlusion and visual field changing on reprojection and adversarial loss, Zhao et al. \cite{masked_gan} introduce the concept of masked GAN. Nevertheless, the adversarial training strategies inherent to these methods frequently lead to compromised network stability. With the design of a new generative network, the diffusion model \cite{ddpm} exhibits better model stability. Song et al. \cite{ddim} develop a more efficient denoising diffusion implicit model (DDIM) to achieve a more reasonable inference time. There have been methods that demonstrated the significant potential of the diffusion model in the realm of supervised depth estimation for enhancing robustness and stability \cite{dif_dep_u, dif_dep}.

\section{Method}
The unsupervised monocular depth estimation framework utilizes geometric constraints from video sequences as the supervision (Section III-A). To enhance the model robustness, we draw inspiration from the diffusion model. We regard the depth estimation task as a denoising process guided by images, which iteratively refines the depth feature distribution (Section III-B). During the denoising process, we propose an innovative hierarchical feature-guided denoising model, enabling the model to learn and interpret the depth feature distribution more effectively (Section III-C). Furthermore, we explore the implicit depth information during reprojection and design a novel implicit depth consistency loss, thereby enhancing the model performance (Section III-D).

\subsection{Background}
In the framework of unsupervised monocular depth estimation, the input of the depth estimation subnetwork is the image at time $t$, denoted as the target image $\boldsymbol{I}_t$. Its output is the depth of the target image, denoted as $\boldsymbol{d}_t$ (Section III-B). Simultaneously, the adjacent frame is served as the source image $\boldsymbol{I}_s$. Both the target image and the source image are fed into the pose estimation subnetwork to obtain the relative pose $\boldsymbol{P}_{t\rightarrow s}$ between these two images. Based on the output of these two subnetworks, we can reproject the source image onto the target image, resulting in the reconstructed target image $\boldsymbol{I}_t'$. Follow Bian et al. \cite{sc-sfm}, the optimization of the aforementioned two subnetworks is achieved by the reprojected photometric loss $L_{ph}$ between $\boldsymbol{I}_t'$ and $\boldsymbol{I}_t$. In addition, to enhance the optimization of the depth within untextured regions of the image, we integrate an edge-aware smoothing loss $L_{sm}$. Furthermore, we incorporate the DDIM loss $L_{ddim}$ and implicit depth consistency loss $L_{dc}$. These two losses will be detailed in Sections III-C and III-D respectively. The comprehensive loss function can be formulated as follows:
$$
L = w_1\cdot L_{ph} + w_2\cdot L_{sm} + w_3\cdot L_{ddim} + w_4\cdot L_{dc}
\eqno{(1)}
$$
where $w_1$ to $w_4$ denote the weights assigned to various losses. The framework of our model is illustrated in Fig.\ref{framework}.

\subsection{Diffusion Depth Network}
In discriminative-based depth estimation methods, the image features $\boldsymbol{F}$ extracted through image feature extraction are directly fed into the depth decoder to obtain its depth $\boldsymbol{x}$. The prediction of depth can be understood as the conditional probability $P(\boldsymbol{x}|\boldsymbol{F})$. During the training process, the network learns to interpret the mapping from $\boldsymbol{F}$ to $\boldsymbol{x}$, expressed as $f(\boldsymbol{F})=\boldsymbol{x}$, where $f$ represents the depth estimation model. However, this learning approach may result in considerable prediction errors when faced with image perturbations. This occurs as biases within the image features $\boldsymbol{F}$ have a direct effect on the mapping to its depth $\boldsymbol{x}$, leading to inaccurate depth estimation and limited robustness.

Unlike discriminative-based methods that directly feed image features into the depth decoder for depth estimation, our proposed diffusion depth network use image features to guide a random distribution through a stepwise denoising process, aiming to generate depth features. The depth features are then fed into the depth decoder to obtain its depth, as shown in Fig.\ref{DDN}. During the denoising process, each step is accomplished by learning the conditional joint probability distribution $p_\theta(\boldsymbol{x}_{n-1}|\boldsymbol{x}_n, \boldsymbol{F})$. This implies that the network is trained to understand the inherent structure and distribution of the depth features, thereby enhancing the network's robustness. Even in the presence of disturbances to the input image, the diffusion depth network can effectively reduce errors arised from biases in the image features.

The diffusion model comprises two processes: the diffusion process and the denoising process. Within the diffusion process, noise is progressively added to the initial distribution $\boldsymbol{x}_0$ to produce the $n_{th}$ distribution $\boldsymbol{x}_n$ through iterative steps. This process plays a crucial role in the DDIM Loss (Section III-C). The diffusion process $q(\boldsymbol{x}_n|\boldsymbol{x}_0)$ is shown in Eq.2:
$$
q(\boldsymbol{x}_n|\boldsymbol{x}_0)=\mathcal{N} (\boldsymbol{x}_n|\sqrt{\overline{\alpha}_n} \boldsymbol{x}_0,(1-\overline{\alpha}_n)\boldsymbol{I})
\eqno{(2)}
$$
where $n\in \{0,1,...,N\}$ represents the diffusion step, $\overline{\alpha}_n=\prod_{s=0}^{n}\alpha_s$, which $\alpha_s$ is the noise variance schedule.

The denoising process aims to remove noise from $\boldsymbol{x}_n$ in order to obtain $\boldsymbol{x}_{n-1}$ through the use of a neural network $\mu_\theta$. The formula for the denoising process can be defined as:
$$
p_\theta(\boldsymbol{x}_{n-1}|\boldsymbol{x}_n, \boldsymbol{F})=\mathcal{N} (\boldsymbol{x}_{n-1}|\mu_\theta(\boldsymbol{x}_n, n, \boldsymbol{F}),\sigma_n^2 \boldsymbol{I})
\eqno{(3)}
$$
where $\sigma_n^2$ denotes the transition variance. To accelerate the denoising process, we utilize the denoising diffusion implicit models (DDIM) \cite{ddim} by setting the variance $\sigma_n^2$ to 0.

\begin{figure}[t]\centering
    \includegraphics[width=0.47\textwidth]{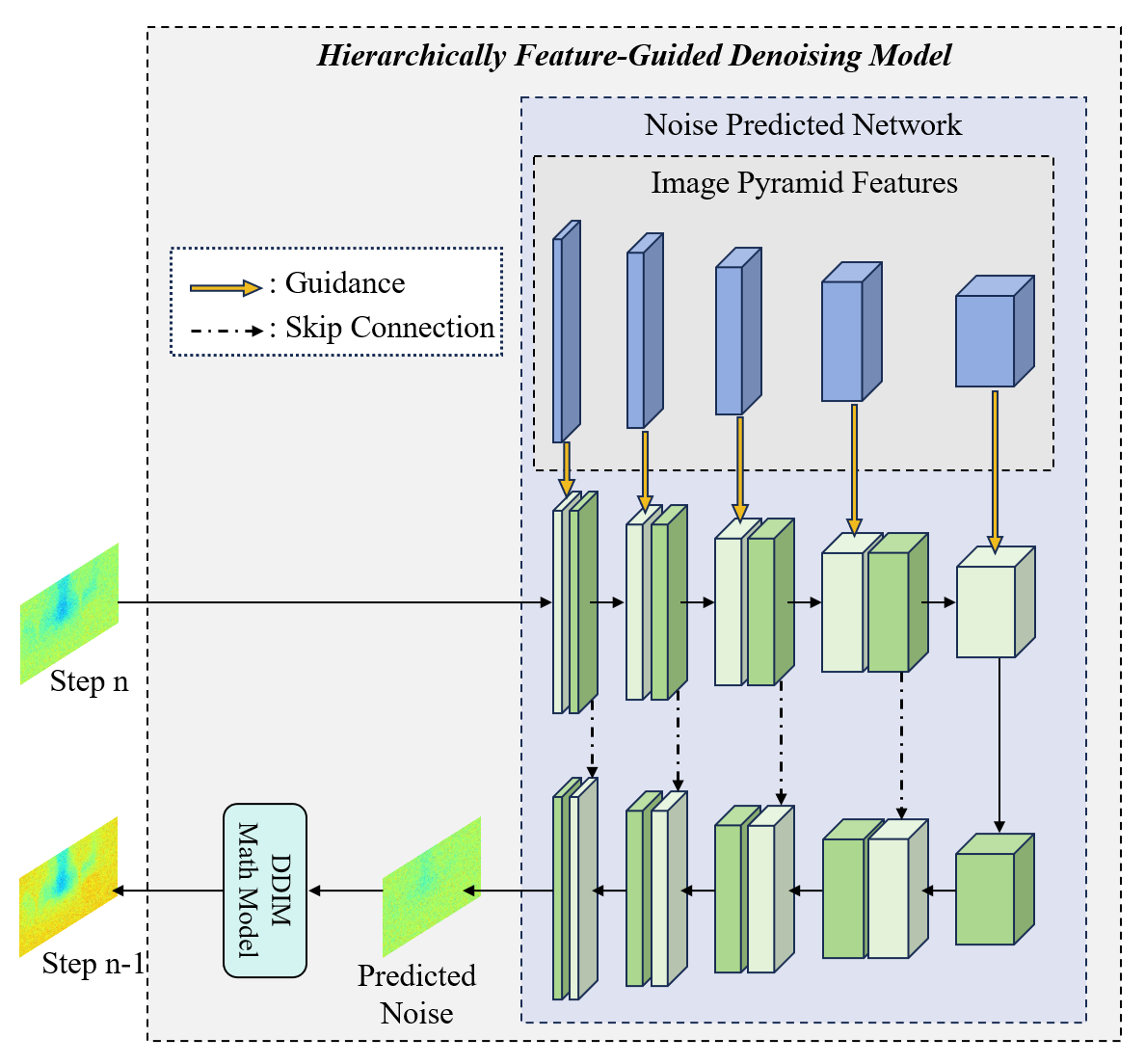}
    \caption{Illustration of hierarchical feature-guided denoising module. We input the distribution at step n, denoted as $\boldsymbol{x}_n$, and utilize image pyramid features to guide its denoising process. The output is denoted as $\boldsymbol{x}_{n-1}$.}
    \label{HFGD}
\end{figure}

\subsection{Hierarchical Feature-Guided Denoising Module}
The hierarchical feature-guided denoising module comprises a noise prediction network $\mu_\theta$ and a DDIM math model, as illustrated in Fig.\ref{HFGD}. Within the diffusion model, the denoising module assumes a crucial role as it is responsible for progressive denoising the initial random distribution $\boldsymbol{x}_N$ during the denoising process. Its task is to take $\boldsymbol{x}_n$ as the input of the noise prediction network, predicting its noise relative to $\boldsymbol{x}_0$. Afterward, $\boldsymbol{x}_{n-1}$ is obtained through the DDIM inference process. Considering the correlation between depth and image, we incorporate the image to guide the denoising process.

In prior research, Saxena et al. \cite{dif_dep_u} employ a direct input of both images and random distributions, utilizing the original RGB images as direct guidance. While the original RGB images do contain color and texture information, they struggle to effectively extract the abundant information in images, such as geometric correlation and semantic informations. Consequently, it is unable to fully utilize the image information as guidance during the denoising process. Extending upon this, Duan et al. \cite{dif_dep} adopt a different approach by aggregating features from the image feature extraction network and integrate them into the middle layer of the denoising module. While this method enriches the guiding features by using the extraction network, the aggregate process confuses spatial geometric information with semantic information, thus not fully capitalizing on the guiding potential of image features. To address this, we specifically propose a hierarchical feature-guided denoising module (HFGD) by guiding the denoising module at various layers with image features of diverse dimensions. This approach fully utilizes the capabilities of image pyramid features for guidance, thus enhance the model's interpretation of depth feature distribution. The framework of HFGD is illustrated in Fig.\ref{HFGD}.

Given that image pyramid features encompass diverse dimensional information, we progressively guide the noise prediction from shallow to deep in HFGD. At the initial prediction stages, we utilize shallow spatial geometric features of the image for guidance. As the network goes deeper, it gains the ability to learn more complex features. Furthermore, our guidance information evolves from low-level spatial geometric features to high-level semantic features, fully capitalizing on the advantages of hierarchical features. This comprehensive utilization enables the model to learn a more refined depth feature distribution. Simultaneously, the diffusion steps $n$ are also embedded and participate in the denoising process as the guidance alongside the image features. Inspired by the U-Net \cite{u_net} architecture, we incorporate skip connections, which allows the noise prediction network to access more high-resolution information during upsampling, enabling better restoration of detailed information.

At the same time, we enhance the model by incorporating the DDIM loss, which is built from the noise consistency in both the diffusing and denoising processes. This loss further constrains the noise prediction network within the model, improving the quality of the generated depth features. By randomly generating noise $\boldsymbol{\epsilon}$ and diffusion step $n'$, we diffuse the output $\boldsymbol{x}_0$ up to the $n'_{th}$ step to obtain $\boldsymbol{x}_n'$ following the procedure defined in Eq.2. Subsequently, $\boldsymbol{x}_n'$, $n'$ and image features $\boldsymbol{F}$ are fed into the noise prediction network $\mu_\theta$ to predict the noise. In theory, the predicted noise and $\boldsymbol{\epsilon}$ should be consistent. The DDIM loss is defined as follows:
$$
L_{DDIM}=\left\|\mu_\theta(\boldsymbol{x}_n', n', \boldsymbol{F}) -  \boldsymbol{\epsilon}\right\|_2
\eqno{(4)}
$$

\begin{table*}[t]
    \center
    \caption{Quantitative results of depth estimation on KITTI raw dataset for distance up to 80m.}
    \label{re_de}
    \begin{tabular}{cc|cccc|ccc}
        \hline
        \multicolumn{2}{c|}{\multirow{2}{*}{Methods}}  & \multicolumn{4}{c|}{Error $\downarrow$}          & \multicolumn{3}{c}{Accuracy $\uparrow$} \\ \cline{3-9} 
                                                                     &                  & Abs Rel & Sq Rel & RMSE  & RMSE log & $\delta<1.25$ & $\delta<1.25^2$ & $\delta<1.25^3$ \\ \hline
        \multicolumn{1}{l|}{\multirow{7}{*}{Discriminative-based}}        & SFMLearner \cite{sfmlearner}       & 0.208   & 1.768  & 6.856 & 0.283    & 0.678    & 0.885   & 0.957   \\
        \multicolumn{1}{l|}{}                                        & SC-SFMLearner \cite{sc-sfm}    & 0.137   & 1.089  & 5.439 & 0.217    & 0.830    & 0.942   & 0.975   \\
        \multicolumn{1}{l|}{}                                        & MonoDepth2 \cite{monodepth2}       & 0.115   & 0.903  & 4.863 & 0.193    & \textbf{0.877}    & 0.959   & 0.981 \\
        \multicolumn{1}{l|}{}                                        & Xiong, et al. \cite{Xiong} & 0.126   & 0.902  & 5.052 & 0.205    & 0.851    & 0.950   & 0.979   \\
        \multicolumn{1}{l|}{}                                        & SC-Depth \cite{sc-depth}         & 0.119   & 0.857  & 4.950 & 0.197    & 0.863    & 0.957   & 0.981   \\
        \multicolumn{1}{l|}{}                                        & VDN \cite{VDN}    & 0.117   & 0.882  & 4.815 & 0.195    & 0.873    & 0.959   & 0.981   \\
        \multicolumn{1}{l|}{}                                        & MonoProb \cite{MonoProb}    & \textbf{0.114}   & 0.861  & 4.765 & 0.190    & 0.876    & \textbf{0.961}   & 0.982   \\ \hline
        \multicolumn{1}{c|}{\multirow{5}{*}{Generative-Based}} & GAN-VO \cite{gan_vo}           & 0.150   & 1.141  & 5.448 & 0.216    & 0.808    & 0.939   & 0.975   \\
        \multicolumn{1}{c|}{}                                        & Li, et al. \cite{gan_vo2}        & 0.150   & 1.127  & 5.564 & 0.229    & 0.832    & 0.936   & 0.974   \\
        \multicolumn{1}{c|}{}                                        & Zhao, et al. \cite{masked_gan}        & 0.139   & 1.034  & 5.264 & 0.214    & 0.821    & 0.942   & 0.978   \\
        \multicolumn{1}{c|}{}                                        & Xu, et al. \cite{mc_gan} & 0.144   & 1.148  & 5.632 & 0.234    & 0.795    & 0.927   & 0.971   \\
        \multicolumn{1}{c|}{}                                        & SharinGAN \cite{SharinGAN} & 0.116   & 0.939  & 5.068 & 0.203    & 0.850    & 0.948   & 0.978   \\
        \multicolumn{1}{c|}{}                                        & \textbf{Ours}     & \textbf{0.114}   & \textbf{0.747}  & \textbf{4.724} & \textbf{0.187}    & 0.863    & 0.960   & \textbf{0.984}   \\ \hline
    \end{tabular}
\end{table*}

\subsection{Implicit Depth Consistency Loss}
During the reprojection process, we can calculate the reprojected depth of the source image by utilizing the depth of the target image $\boldsymbol{d}_t$ and the pose between the target and source images $\boldsymbol{P}_{t\rightarrow s}$. We aim to use this reprojected depth as an implicit pseudo-label to better constrain the depth estimation subnetwork to enhance its performance. Since the computation involves the network-estimated depth of the target image, this constraint also helps to ensure that depth estimation within a monocular video sequence remains consistent in scale.

Since Bian et al. \cite{sc-sfm, sc-depth} employ the reprojected depth to construct a geometric consistency loss, which is calculated through the difference between the network-estimated depth of the source image and the reprojected depth. We acknowledge that the mask constructed using this loss plays a crucial role in filtering dynamic objects. However, due to the relatively lower accuracy of estimated poses during the initial stages of training, reprojection can easily result in erroneous correspondences. This can lead to differences between the reprojected depth and the network-estimated depth which may be caused by incorrect correspondences rather than depth estimation errors.

For this reason, we design an improved approach by proposing the implicit depth consistency loss. During the reprojection, we obtain correspondence information between the target and source images. In the reprojected photometric loss, where we reproject the source image onto the target image to generate the reconstructed target image. We can similarly reproject the target image onto the source image based on the correspondence information to obtain the reconstructed source image $\boldsymbol{I}_s'$. The reconstructed source image $\boldsymbol{I}_s'$ is then passed through the depth estimation subnetwork to generate the network-estimated reconstructed source depth, as depicted in Fig.\ref{framework}. Since the reprojected depth and the network-estimated reconstructed source depth do not encounter the issue of incorrect correspondences, they are theoretically expected to be the same. We formulate the implicit depth consistency loss $L_{dc}$ as follows:
$$
L_{dc}=\left\|\boldsymbol{I}_s^{-1}\cdot \boldsymbol{K} \boldsymbol{P}_{t\rightarrow s}(\boldsymbol{d}_t \cdot \boldsymbol{K}^{-1}\boldsymbol{I}_t) - {\rm DDN}(\boldsymbol{I}_s') \right\|_1
\eqno{(5)}
$$
where $\boldsymbol{K}$ denotes the camera intrinsics, and ${\rm DDN}$ represents the diffusion depth network.

\section{Experiment}
\subsection{Dataset}
\textbf{KITTI.} The KITTI dataset \cite{kitti} is currently the most widely used benchmark dataset for evaluating computer vision algorithms in the context of autonomous driving due to its wide variety of sensor data and realistic scenarios. For depth evaluation, we partition the KITTI raw dataset using Eigen's split method \cite{eigen} with 39,810 training, 4,424 validation, and 697 test images.

\textbf{Make3D.} The Make3D dataset \cite{make3d} comprises a collection of images captured from a variety of scenes, each accompanied by its corresponding depth map. The dataset encompasses a total of 534 images, with 400 images for training and 134 images for testing. Given the relatively small size of the training set, this dataset is predominantly utilized for evaluating generalization capabilities.

\begin{figure}[t]\centering
    \includegraphics[width=0.47\textwidth]{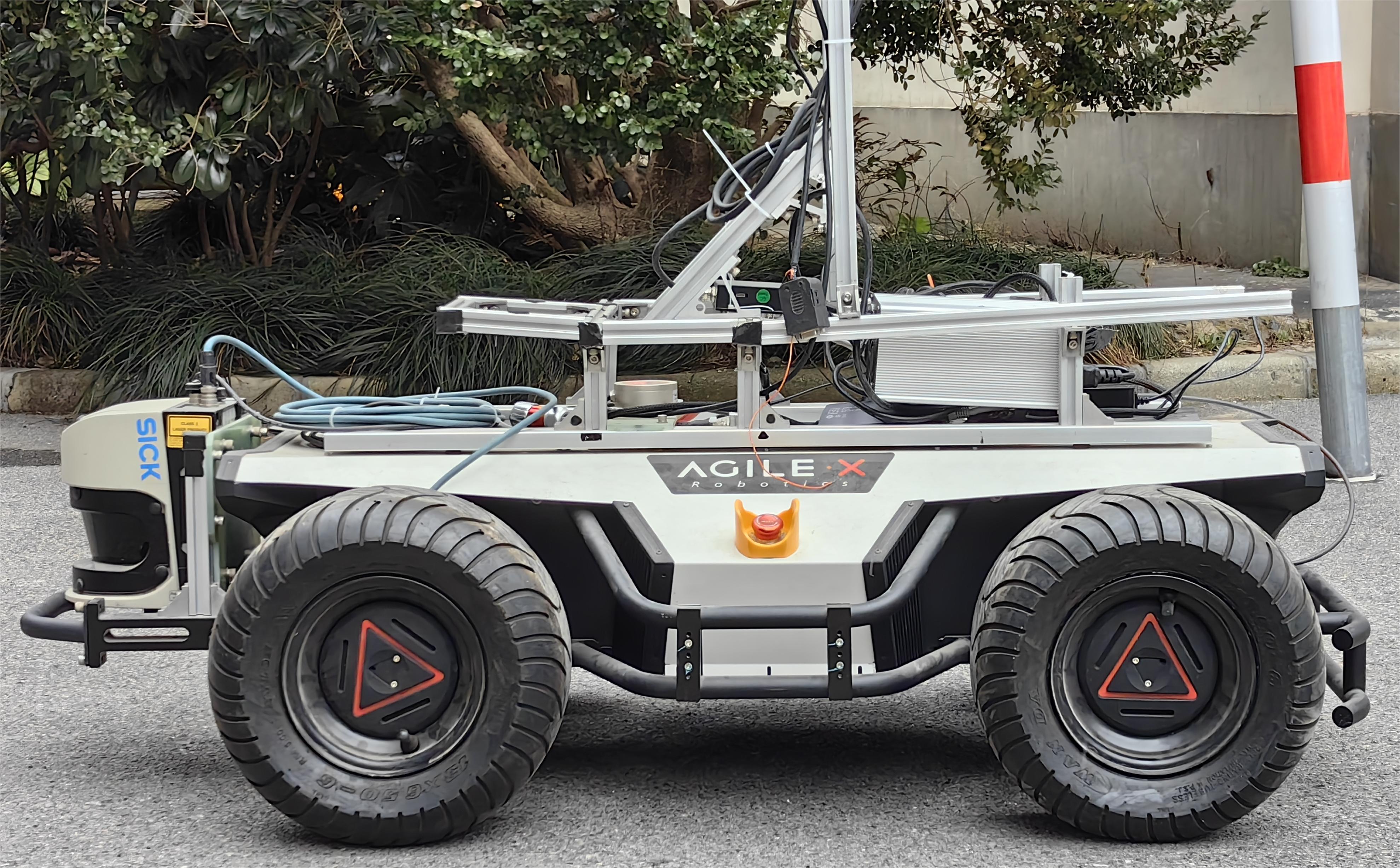}
    \caption{The mobile robot we use to collect the SIMIT dataset.}
    \label{simit_car}
\end{figure}

\textbf{SIMIT.} The SIMIT dataset comprises images we collected from outdoor environments. We use the mobile robot shown in Figure \ref{simit_car} to capture scenes of the nearby streets, which include the sky, trees, pedestrians, vehicles, and more. We use this self-collected dataset to evaluate the generalizability of different methods.

\subsection{Implementation Details}
The proposed method is implemented using the PyTorch library. We employ the Adam optimizer and set the learning rate to $10^{-4}$. We employ ResNet-18 \cite{resnet} which is pretrained on ImageNet \cite{imagenet} to extract image features in the diffusion depth network. The pose estimation subnetwork consists of a ResNet-18 encoder and two fully connected layers which is the same as SC-Depth \cite{sc-depth}. Considering the initial low accuracy of depth generated during the early training, we affiliate DDIM loss $L_{ddim}$ at the $20_{th}$ epoch. We adhere to the training strategy outlined in SC-Depth \cite{sc-depth}, using sequences of three consecutive video frames as training samples. We calculate projections and losses from the second frame to the other frames and reverse them again to maximize data utilization. During the training process, images are enhanced through random scaling, cropping, and horizontal flipping. In Eq.1, the value of $w_1$ is set to 1.0, while $w_2$ to $w_4$ are assigned a value of 0.1 each. Following Bian et al. \cite{sc-sfm, sc-depth}, we convert the sigmoid output of the depth estimation subnetwork to depth with $\boldsymbol{D} = 1/(a\boldsymbol{x} + b)$, where $a$ is equal to 10 and $b$ is equal to 0.01.

\begin{table*}[t]
    \center
    \caption{Quantitative results of depth estimation on KITTI raw dataset in challenging scenarios.}
    \label{re_de_ro}
    \begin{tabular}{cc|cccc|ccc}
    \hline
    \multicolumn{1}{c|}{\multirow{2}{*}{Conditions}} & \multirow{2}{*}{Methods}    & \multicolumn{4}{c|}{Error $\downarrow$}            & \multicolumn{3}{c}{Accuracy $\uparrow$}                     \\ \cline{3-9} 
    \multicolumn{1}{c|}{} &              & Abs Rel        & Sq Rel         & RMSE           & RMSE log       & $\delta<1.25$ & $\delta<1.25^2$ & $\delta<1.25^3$  \\ \hline
    \multicolumn{1}{c|}{\multirow{4}{*}{Motion Blur}} & MonoDepth2 \cite{monodepth2}  & 0.162          & 1.308          & 6.148          & 0.257          & 0.774          & 0.914          & 0.960          \\
    \multicolumn{1}{c|}{}                      & SC-Depth \cite{sc-depth}     & 0.182          & 1.509          & 6.824          & 0.285          & 0.724          & 0.891          & 0.949          \\
    \multicolumn{1}{c|}{}                      & MonoProb \cite{MonoProb}     & 0.190          & 1.643          & 6.612          & 0.288          & 0.724          & 0.887          & 0.947          \\
    \multicolumn{1}{c|}{}                      & \textbf{Ours} & \textbf{0.144} & \textbf{1.062} & \textbf{5.905} & \textbf{0.231} & \textbf{0.794} & \textbf{0.930} & \textbf{0.973} \\ \hline
    \multicolumn{1}{c|}{\multirow{4}{*}{Rainy}} & MonoDepth2 \cite{monodepth2}  & 0.257          & 2.488          & 7.300          & 0.349          & 0.591          & 0.830          & 0.922          \\
    \multicolumn{1}{c|}{}                      & SC-Depth \cite{sc-depth}     & 0.250          & 2.215          & 7.407          & 0.347          & 0.593          & 0.832          & 0.926          \\
    \multicolumn{1}{c|}{}                      & MonoProb \cite{MonoProb}     & 0.252          & 2.357          & 7.316          & 0.341          & 0.598          & 0.838          & 0.930 \\
    \multicolumn{1}{c|}{}                      & \textbf{Ours} & \textbf{0.208} & \textbf{1.568} & \textbf{6.387} & \textbf{0.287} & \textbf{0.665} & \textbf{0.885} & \textbf{0.955} \\ \hline
    \multicolumn{1}{c|}{\multirow{4}{*}{Presence of Noise}}  & MonoDepth2 \cite{monodepth2}  & 0.143          & 1.150          & 5.348          & 0.223          & 0.817          & 0.941          & 0.975          \\
    \multicolumn{1}{c|}{}                      & SC-Depth \cite{sc-depth}     & 0.141          & 1.028          & 5.435          & 0.223          & 0.807          & 0.937          & 0.975          \\
    \multicolumn{1}{c|}{}                      & MonoProb \cite{MonoProb}     & 0.144          & 1.120          & 5.305          & 0.222          & 0.813          & 0.940          & 0.975 \\
    \multicolumn{1}{c|}{}                      & \textbf{Ours} & \textbf{0.130} & \textbf{0.841} & \textbf{4.948} & \textbf{0.203} & \textbf{0.833} & \textbf{0.951} & \textbf{0.982} \\ \hline \hline
    \multicolumn{1}{c|}{\multirow{4}{*}{Average}}  & MonoDepth2 \cite{monodepth2}  & 0.187          & 1.649          & 6.265          & 0.276          & 0.727          & 0.895          & 0.952          \\
    \multicolumn{1}{c|}{}                      & SC-Depth \cite{sc-depth}     & 0.191          & 1.584          & 6.555          & 0.285          & 0.708          & 0.887          & 0.950          \\
    \multicolumn{1}{c|}{}                      & MonoProb \cite{MonoProb}     & 0.195          & 1.707          & 6.411          & 0.284          & 0.712          & 0.888          & 0.951 \\
    \multicolumn{1}{c|}{}                      & \textbf{Ours} & \textbf{0.161} & \textbf{1.157} & \textbf{5.746} & \textbf{0.240} & \textbf{0.764} & \textbf{0.922} & \textbf{0.970} \\ \hline
    \end{tabular}
\end{table*}

\begin{figure*}[!t]\centering
    \includegraphics[width=0.9\textwidth]{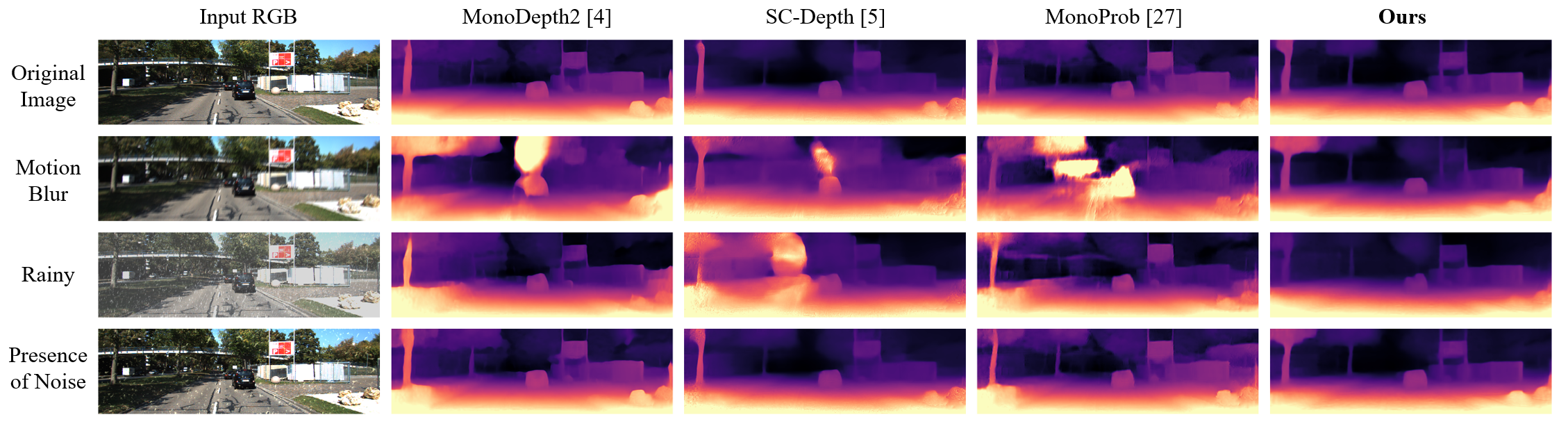}
    \includegraphics[width=0.9\textwidth]{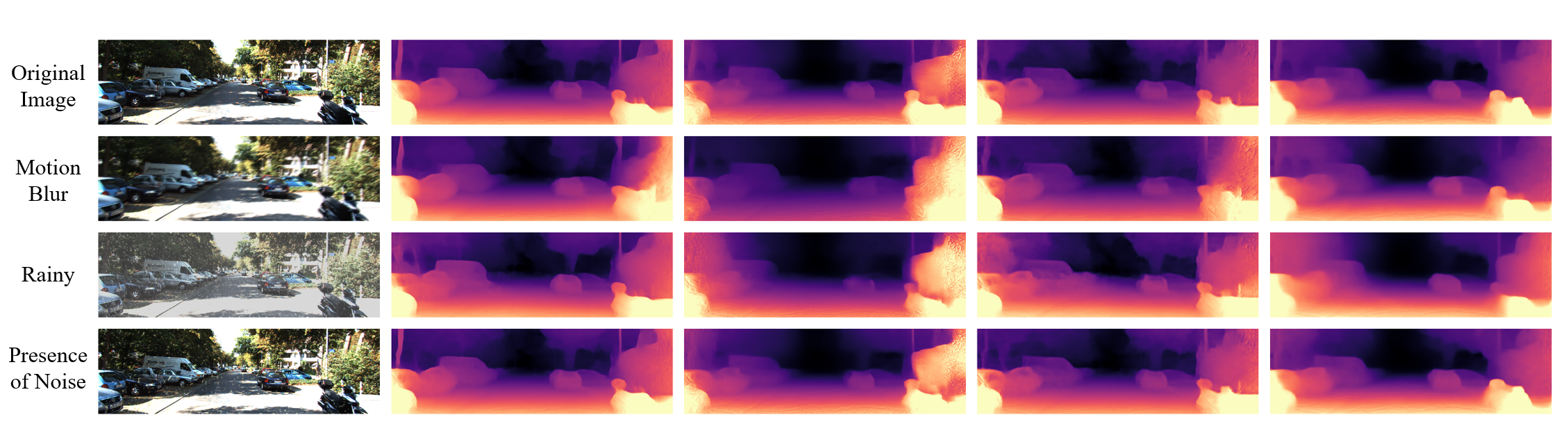}
    \includegraphics[width=0.9\textwidth]{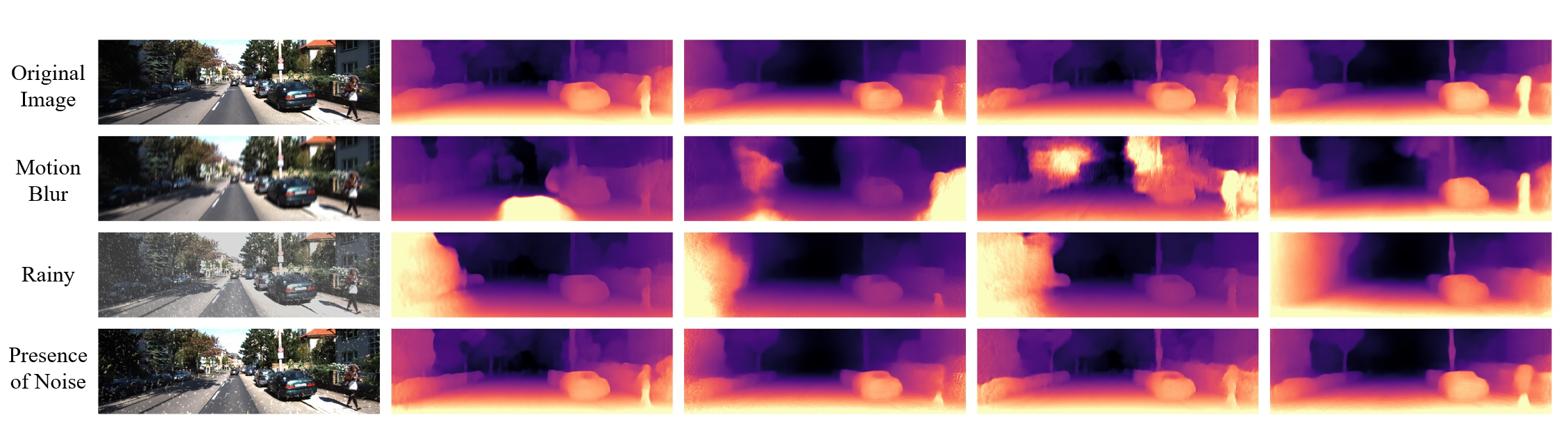}
    \caption{Qualitative comparison on the KITTI raw dataset and different types of simulated images.}
    \label{result_k}
\end{figure*}

\begin{figure*}[!t]\centering
    \includegraphics[width=0.9\textwidth]{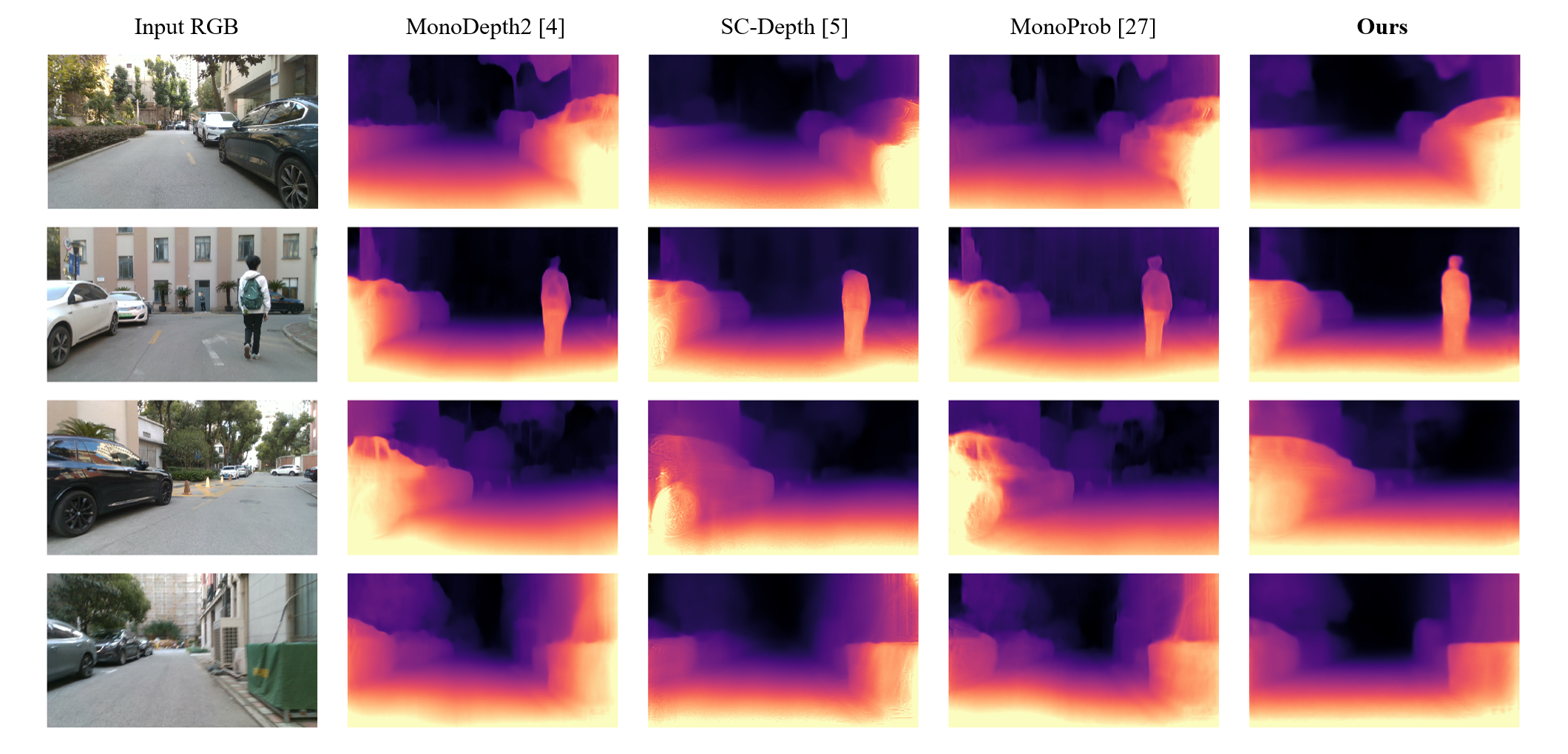}
    \caption{Qualitative comparison of the generalization capabilities on the SIMIT dataset.}
    \label{result_s}
\end{figure*}

\subsection{Main Results}
We first evaluate the depth predicted by our method on the KITTI raw dataset using the metrics described in \cite{eigen}, as shown in Table \ref{re_de}.
We show that our proposed method outperforms among generative-based methods.
Due to different training strategies, discriminative networks directly learn the mapping between input and output, while generative networks aim to learn the distribution of data. Although the accuracy of generative-based methods is slightly less compared to discriminative-based methods, generative-based methods demonstrate greater robustness. We compare our algorithm with several typical monocular depth estimation methods based on discriminative networks. Our method exhibits a comparable level of these methods.

In the test set of the KITTI raw dataset, the images used for evaluation are ideally clear and of high quality. However, in real-world driving scenarios, captured images could be affected by factors such as camera shake, weather, etc., leading to blurry or noisy images. Hence, to evaluate the robustness of our methods, we apply the Imgaug library \cite{imgaug} to process the test set of KITTI raw, generating simulated images with motion blur, rainy conditions, and the presence of noise on the camera, which emulates the scenario where irregular dew has attached to the camera sensors during the early morning. According to the categorization of anomalies in autonomous driving \cite{auto_drive_challenge}, images with motion blur and rainy conditions can be associated with domain-level anomalies, while the presence of noise on the camera sensors can be associated with pixel-level anomalies. Both types of anomalies are relatively common in real-world driving scenarios.

To verify the robustness of our method, we conducted tests in the three aforementioned scenarios and compared it with several methods. The robustness evaluation results are summarized in Table \ref{re_de_ro} and then illustrate their performance qualitatively in Figure \ref{result_k}. It can be seen from Figure \ref{result_k} that our method exhibits no significant deviation in both ideal test sets and those subjected to perturbations, in contrast to several other methods that display considerable biases. In Table \ref{re_de_ro}, the first three parts in the table represent the three types of scenarios, and the fourth part represents the average error and accuracy. It is evident that our method outperforms the other methods across all evaluation metrics, demonstrating its strong robustness.

\begin{table}[t]
    \center
    \caption{Quantitative results of depth estimation on Make3D dataset.}
    \label{re_make3d}
    \resizebox{0.47\textwidth}{!}{
    \begin{tabular}{c|cccc}
    \hline
    \multirow{2}{*}{Methods} & \multicolumn{4}{c}{Error $\downarrow$}           \\ \cline{2-5} 
                             & Abs Rel & Sq Rel & RMSE  & RMSE log \\ \hline
    SFMLearner \cite{sc-sfm}               & 0.383   & 5.321  & 10.47 & 0.478    \\
    Xiong, et al. \cite{Xiong}  & 0.320   & 3.170  & 7.062 & 0.163   \\
    Monodepth2 \cite{monodepth2}               & 0.322   & 3.589  & 7.417 & 0.163    \\
    SC-Depth \cite{sc-depth}               & 0.362   & 3.927  & 7.768 & 0.180    \\
    MonoProb \cite{MonoProb}                & 0.327   & -      & \textbf{6.687} & -        \\ \hline
    Zhao, et al. \cite{masked_gan} & 0.312   & 2.914  & 6.863 & 0.163        \\
    SharinGAN \cite{SharinGAN}                & 0.377   & 4.900  & 8.388 & -        \\
    \textbf{Ours}                     & \textbf{0.295}   & \textbf{2.633}  & 7.103 & \textbf{0.162}    \\ \hline
    \end{tabular}}
\end{table}

We evaluate our method on the Make3D and SIMIT datasets to show its generalization ability in different outdoor scenes. We use the model trained on the KITTI raw dataset without any fine-tuning. Table \ref{re_make3d} show the comparison of our method with the other methods on the Make3D dataset. The upper part is methods base on discriminative network and the half bottom is methods base on generative network. It can be seen that our method has the smallest absolute relative and square relative error. This shows that the generalizability of our method is considerable. Figure \ref{result_s} shows the qualitative analysis of our method on the SIMIT dataset. Affected by the turbulence of the mobile robot when collecting images, most of the images we collected are blurry. Our method performs well in such blurry images, especially the depth of foreground objects in the scene. This not only demonstrates the excellent generalization capabilities of our method but also underscores its significant robustness.

\begin{table}[t]
    \center
    \caption{Ablation study of HFGD and $L_{dc}$ on the KITTI dataset.}
    \label{ab_hfgd}
    \begin{tabular}{cc|cccc}
    \hline
    \multicolumn{1}{c}{\multirow{2}{*}{HFGD}} & \multirow{2}{*}{$L_{dc}$} & \multicolumn{4}{c}{Error}           \\ \cline{3-6} 
    \multicolumn{1}{c}{}                      &                    & Abs Rel & Sq Rel & RMSE  & RMSE log \\ \hline
                                              &                    & 0.123   & 0.893  & 4.915 & 0.195    \\
                                              &      $\bigstar$    & 0.121   & 0.844  & 4.862 & 0.193    \\
                            $\bigstar$        &                    & 0.115   & 0.792  & 4.749 & 0.190    \\
                            $\bigstar$        &      $\bigstar$    & \textbf{0.114}   & \textbf{0.747}  & \textbf{4.724} & \textbf{0.187}        \\ \hline
    \end{tabular}
    \end{table}

\subsection{Ablation Studies}
To verify the effectiveness of our proposed HFGD and implicit depth consistency loss $L_{dc}$, we conduct ablation experiments as shown in Table \ref{ab_hfgd}. We designate the method that does not incorporate HFGD and $L_{dc}$ as the base model. When not utilizing HFGD, we follow the approach in \cite{dif_dep} to fuse image features and input them into the intermediate layer of the network to guide the denoising process. Building upon the base model, we integrate an implicit depth consistency loss $L_{dc}$, as shown in the second row of the table. The results indicate that the inclusion of $L_{dc}$ enhances the model's performance by more effectively constraining the depth estimation subnetwork. Then, we replaced the image guidance approach in the base model with HFGD, as depicted in the third row of the table. It is evident that the model's accuracy has significantly improved. This demonstrates that HFGD's progressive guidance approach, from low-level spatial geometric information to high-level semantic information, can more effectively enable the model to learn and interpret the distribution of depth features. As a result, it enhances the precision of the model's depth estimation. Finally, the results show that the model performs better when using HFGD and implicit depth consistency loss.

\section{Conclusion}
This paper proposes an unsupervised monocular depth estimation method based on the diffusion model. Benefiting from the generative network, the diffusion model, our method exhibits strong robustness. It performs exceptionally well in two common challenging scenarios encountered in autonomous driving, domain-level and pixel-level anomalies. We improve the guidance approach for image features during the denoising process by utilizing a hierarchical feature-guided denoising module. This approach allows for a more comprehensive utilization of both spatial geometry and semantic features from the image, thus enabling the model to learn the enhanced depth feature distribution. Furthermore, we design a novel implicit depth consistency loss, which provides the depth estimation subnetwork with additional constraint. It enhances our model's performance and makes sure the estimated depths are consistent in scale within the same video sequence. Experimental results show that our approach achieves promising estimation and remarkable robustness, which is particularly useful in real-world scenarios.

\end{document}